\documentclass[manuscript,screen]{acmart}

\AtBeginDocument{%
  }

\usepackage{graphicx}
\usepackage{caption}
\usepackage{subcaption}
\usepackage{booktabs}

\usepackage{adjustbox}
\usepackage{rotating}
\usepackage{booktabs}
\usepackage{multirow}
\usepackage{longtable}

\usepackage{hyperref}
\usepackage{xcolor}
\usepackage[linesnumbered,ruled]{algorithm2e}

\SetCommentSty{mycommfont}
\newcommand{\eg}{{$\epsilon$-greedy }}

\begin{document}

\title{Learning in Cooperative Multiagent Systems Using Cognitive and Machine Models}

\author{Thuy Ngoc Nguyen}
\affiliation{%
  \institution{University of Dayton}
  \streetaddress{300 College Park}
  \city{Dayton}
  \state{Ohio}
  \country{USA}
  \postcode{45469}
}
\email{ngoc.nguyen@udayton.edu}
\author{Duy Nhat Phan}
\email{dnphan@andrew.cmu.edu}

\author{Cleotilde Gonzalez}

\email{coty@cmu.edu}
\affiliation{%
  \institution{Dynamic Decision Making Laboratory, Carnegie Mellon University}
  \streetaddress{5000 Forbes Ave.}
  \city{Pittsburgh}
  \state{Pennsylvania}
  \country{USA}
  \postcode{15213}
}

% \affiliation{%

% \affiliation{%

% \affiliation{%

% \affiliation{%

% \affiliation{%

% \affiliation{%

% \affiliation{%

\renewcommand{\shortauthors}{Nguyen et al.}

\begin{abstract}
% Developing effective Multi-Agent Systems (MAS) is critical for many applications requiring collaboration and coordination. Despite the advance of deep reinforcement learning (MADRL) in cooperative MAS, one of the major challenges remaining is the simultaneous
% learning and interaction of independent agents in dynamic environments with the presence of stochastic rewards. State-of-the-art MADRL models struggle to perform well in Coordinated Multi-agent Object Transportation Problems (CMOTPs) wherein agents must coordinate with each other and learn from stochastic rewards. In contrast, humans often learn rapidly to adapt to nonstationary environments that require coordination among people. In this paper, motivated by the demonstrated ability of cognitive models based on Instance-Based Learning Theory (IBLT) that can capture human decisions in many dynamic decision making tasks, we propose three variants of Multi-Agent IBL models (MAIBL). The idea of these MAIBL algorithms is to combine the cognitive mechanisms of IBLT and the techniques of MADRL models to deal with coordination MAS in stochastic environments from the perspective of independent learners. We demonstrate that the MAIBL models exhibit faster learning and better coordination in a dynamic CMOTP task with various settings of stochastic rewards compared to MADRL models. We discuss the benefits of integrating cognitive insights into MADRL models.

Developing effective Multi-Agent Systems (MAS) is critical for many applications requiring collaboration and coordination with humans.  Despite the rapid advance of Multi-Agent Deep Reinforcement Learning (MADRL) in cooperative MAS, one of the major challenges that remain is the simultaneous learning and interaction of independent agents in dynamic environments in the presence of stochastic rewards. State-of-the-art MADRL models struggle to perform well in Coordinated Multi-agent Object Transportation Problems (CMOTPs), wherein agents must coordinate with each other and learn from stochastic rewards. In contrast, humans often learn rapidly to adapt to nonstationary environments that require coordination among people. In this paper, motivated by the demonstrated ability of cognitive models based on Instance-Based Learning Theory (IBLT) to capture human decisions in many dynamic decision making tasks, we propose three variants of Multi-Agent IBL models (MAIBL). The idea of these MAIBL algorithms is to combine the cognitive mechanisms of IBLT and the techniques of MADRL models to deal with coordination MAS in stochastic environments, from the perspective of independent learners. We demonstrate that the MAIBL models exhibit faster learning and achieve better coordination in a dynamic CMOTP task with various settings of stochastic rewards compared to current MADRL models. We discuss the benefits of integrating cognitive insights into MADRL models.
\end{abstract}

\ccsdesc[500]{Computing methodologies~Multi-agent systems}
\ccsdesc[500]{Computing methodologies~Cognitive science}

\keywords{coordination problems, instance-based learning theory, multi-agent deep reinforcement learning, multi-agent instance-based learning}

\maketitle

\section{Introduction}
Learning to coordinate in cooperative multiagent systems (MAS) has been a central problem that has drawn much attention in interdisciplinary research from robotics,  economics, web-service technology~\cite{wang2017integrating}, and in artificial intelligence (AI) communities~\cite{lauer2000algorithm}.
Coordination, in this context, refers to the ability of two or more agents to jointly reach an agreement on the actions to perform in an environment.
Applications of such coordinating tasks can be found in diverse settings, for instance, a team of robots working together to locate victims in a search and rescue situation~\cite{jennings1997cooperative}, or a group of robots required to coordinate to pick up, carry, and deliver goods in object transportation problems~\cite{rus1995moving}, {\color{black} or in the context of autonomous vehicles, where the coordination between autonomous vehicles and human drivers is crucial \cite{toghi2021altruistic,toghi2022social,toghi2021cooperative}}. This highlights the need to develop MAS that can learn to coordinate and cooperate with humans and other agents.

% %: Incorporate ideas in the Diepold's paper
In the MAS literature, learning coordination is a cooperative problem wherein multiple agents are encouraged to work together towards a common goal by receiving an equally-shared reward~\cite{MatignonLF12}.
Moreover, based on the information available to the agents, they can be independent or joint-action learners~\cite{ClausB98,gronauer2021multi}. The independent learners are essentially non-communicative agents; they have no knowledge of the rewards and actions of the other agents. In contrast, joint-action learners are aware of the existence of other agents and can observe others' actions. Many practical control applications feature a team of multiple agents that must coordinate independently to achieve a common goal without awareness of other members' actions~\cite{AgoginoT12,VerbeeckNPT07}. One example of this problem is a team of rescuers splitting up in a network of underground caves wherein information exchange is unavailable. In the current research, we are interested in the coordination problem from the perspective of independent learners who try to learn and adapt their actions to the other teammates' behavior without communicating during learning.

There are a number of challenges for modeling independent learners in multiagent cooperative tasks. One major problem is \textit{simultaneous learning} within a shared dynamic environment. If the agent selects what appears to be an optimal action for itself as an individual, and if all agents act with the same goal, the situation would result in poor joint actions (i.e., miscoordination). For instance, if a group of drivers departs from one location to the same destination and the driving navigator provides all drivers with the same path, a miscoordination and traffic jam could be generated. Therefore, the navigator's strategies are subject to change over time, and the other agent also has to adapt to this change. Consequently, the presence of other learning and exploring agents make the environment a non-stationary and dynamic situation from the perspective of the independent agents~\cite{tan1993multi}. {\color{black} A number of studies have been proposed to cope with the non-stationarity problem~\cite{foerster2017learning,foerster2017stabilising}, and yet, these works were inspired by centralized learning where agents can communicate freely~\cite{foerster2018counterfactual}, rather than non-communicative agents.}

To address this emergent challenge with non-communicative agents, numerous methods have been proposed in the literature of multiagent reinforcement learning (MARL), including distributed Q-learning \cite{LauerR00}, hysteretic Q-learning \cite{matingon07,MatignonLF12}, and lenient Q-learning \cite{PanaitSL06,WeiL16}. In general, distributed and hysteretic learning operate under an optimistic learning assumption. That is, an agent selects an action that meets the expectation that the other agents also choose the best matching actions accordingly. Under this assumption, the agent prefers positive results when playing actions. Alternatively, the lenient method rests on the assumption that agents are more lenient initially when exploration is still high, but they become less lenient over time. Simply put, the more agents explore the environment, the less lenient they become. This idea is formulated in the model through a parameter (i.e., temperature value) that is used to control the degree of leniency. Particularly, the model associates the selected actions with the temperature value that gradually decays by the frequency of state-action pair visits. 
In another thread of research on how to achieve effective coordination between agents in MAS, \cite{hao2014multiagent,hao2013achieving} proposed a social learning framework that focuses on achieving socially optimal outcomes with reinforcement social learning agents.
Recently, the integration of deep learning into these traditional Reinforcement Learning (RL) methods led to new branches of research, including multi-agent deep reinforcement learning (MADRL) approaches~\cite{OmidshafieiPAHV17,GuptaEK17,PalmerTBS18,LanctotZGLTPSG17}.

% %Stochasticity
Despite the advance in MADRL, these algorithms still perform poorly in tasks in which the environment is not stationary due to the dynamics of coexisting agents and the presence of stochastic rewards. Indeed, the presence of stochastic rewards can add further complications to the cooperation among agents since agents are not always able to distinguish the environment's stochasticity from another agent's exploration~\cite{ClausB98}. Previous research showed that independent learners often result in poor coordination performance in complex or stochastic environments~\cite{ClausB98,lauer2000algorithm,gronauer2021multi}.  This is perhaps due to ambiguity in the source of stochasticity since it can emerge from many factors, including possible outcomes or their likelihood. Despite the fact that prior studies have presented different approaches to cope with stochasticity in MAS~\cite{matingon07,omidshafiei2017deep,PalmerTBS18}, there is significant room left for characterizing sources of stochastic rewards and addressing their effects on the performance of independent agents in the context of fully cooperative MAS.

It is well-known that humans have the ability to adapt to non-stationary environments with stochastic rewards rapidly, and they learn to collaborate and coordinate effectively, while algorithms cannot capture this human ability~\cite{lake2017building}.  Yet, when humans confront stochastic rewards with small-probability outcomes (henceforth referred to as rare events), the decision problem might become complex not only for algorithms but for humans too ~\cite{hertwig2004decisions}. These situations involving highly impacting but rare events are, in fact, very common in real life (e.g., disasters, economic crashes) \cite{taleb2007black}. In decision making, dynamic cognitive models inspired by the psychological and cognitive processes of human decision making have been able to capture and explain the tendency for humans to underweight rare events \cite{GONZALEZ11, hertwig2015}. This current state of affairs motivates the main idea of our paper: constructing algorithms that combine the strengths of MADRL models and cognitive science models, to develop agents that can improve their learning and performance in non-stationary environments with stochastic reward and rare events.

Cognitive modeling has been developed to understand and interpret human behavior by representing the cognitive steps by which a task is performed. In particular, Instance-based Learning Theory (IBLT) was developed to provide a cognitively-plausible account for how humans make decisions from experience and under uncertainty, through interactions with dynamic environments~\cite{GONZALEZ03}. IBLT has shown an accurate representation of human choice and broad applicability in a wide number of decision making domains, from economic decision making to highly applied situations, including complex allocation of resources and cybersecurity, e.g.~\cite{hertwig2015,GONZALEZ2013, GONZALEZ03}. Also, recent work in combining a cognitive model based on IBLT and the temporal difference (TD) mechanism in RL (IBL-TD) has shed light on how to exploit the respective strengths of cognitive IBL and RL models~\cite{Nguyen21}. The idea of IBL-TD and the disadvantages of current MADRL in stochastic rewards, lead to questions of how would the IBL-TD perform in the context of cooperative MAS? {\color{black}And} how would MAS that exploit cognitive models compare to state-of-the-art MADRL approaches to address fully cooperative tasks with stochastic rewards? 

To that end, our first contribution here is to propose novel multi-agent IBL (MAIBL) models that combine the ideas of cognitive IBL models and concepts used in MADRL models, namely decreasing $\epsilon$-greedy, hysteretic, and leniency to solve fully cooperative problems from the perspective of independent agents. Next, we characterize different properties of stochastic rewards, including problems with rare events, to aim at understanding their effects on the behavior of independent agents in fully cooperative Coordinated Multi-Agent Object Transportation Problems (\texttt{CMOTP}), which have been used to test current MADRL algorithms \cite{PalmerTBS18}. Finally, we evaluate the performance of our proposed MAIBL models against these state-of-the-art approaches in the MADRL literature, including decreasing $\epsilon$-greedy, hysteretic, and lenient Deep Q-Network algorithms \cite{PalmerTBS18} on four scenarios of \texttt{CMOTP} with respect to varying stochastic rewards. We demonstrate that MAIBL can significantly outperform MADRL with respect to different evaluation metrics across the choice scenarios.   

\section{Multi-agent deep reinforcement learning} \label{sec:background}
In general, a fully cooperative multi-agent problem is formulated as a Markov game, which is defined by a set of states $\mathcal S$ describing the possible configurations of all agents, a set of actions $\mathcal A_1, ..., \mathcal A_m$, and a set of observations $\mathcal O_1, ..., \mathcal O_m$ for each agent. $\mathbf A = \mathcal A_1\times ...\times\mathcal A_m$ is the joint action set. To select actions, each agent $i$ uses its policy $\pi_i:\mathcal O_i\times\mathcal A_i \to \mathbb R$, which produces the next state according to the transition function $\mathcal T: \mathcal S\times \mathbf A \to \mathcal S$. The agent $i$ receives rewards {\color{black}based on a conditional probability $P_i:\mathcal S\times\mathbf A\times \mathbb R \to [0,1]$ determining the probability of achieving a reward $r\in \mathbb R$ if the joint action $a\in\mathbf A$ has been executed in the current state $s\in\mathcal S$,} and receives private observations correlated with the states $\mathbf O_i:\mathcal S\to \mathcal O_i$.

To date, a large body of research on MADRL focused on building agents that can quickly learn an optimal joint policy in cooperative multi-agent tasks. The most fundamental, as well as recent approaches to MADRL, are summarized in the following.

\subsection{Greedy-MADRL}
Q-learning \cite{WatkinsD92}, one of the most popular single-agent RL algorithms, was among the first algorithms applied to multi-agent settings due to its simplicity and robustness. In the Q-learning algorithm~\cite{WatkinsD92}, $Q_i:\mathcal O_i\times \mathcal A_i\to \mathbb R$ is the Q function of the agent $i$. The Q-value $Q_i(o,a)$ of the observation-action pair $(o,a)$ can be updated by
\begin{equation}
    Q_i(o,a) \longleftarrow Q_i(o,a) + \alpha\delta,
\end{equation}
where $\delta =  r_i +\gamma\max_{a'\in \mathcal A_i}Q_i(o',a') - Q_i(o,a)$ is the Temporal Difference (TD) error with $r_i$ being the reward, $\gamma$ being a discount factor and $o'$ being the observation at the next state $s'$, and $\alpha$ is a learning rate.

Double Deep Q-Network (DQN) \cite{HasseltGS16} approximates the $Q$-value function by minimizing the loss
\begin{equation}\label{lossDQN}
    \mathcal L_i(\theta) = \mathbb E_{o,a,r,o'}\biggl(r + \gamma\bar Q_i(o',a'|\bar\theta) - Q_i(o,a|\theta)\biggr)^2,
\end{equation}
where $\bar Q_i$ is the target Q function of $Q_i$, whose parameters $\bar\theta$ are periodically updated with the most recent $\theta$, which helps stabilize learning, and $a'\in\arg\max_{a''\in \mathcal A_i}Q_i(o',a''|\theta)$. The idea of the double DQN using a decreasing $\epsilon$-greedy exploration strategy (Greedy-MADRL) is that the agent $i$ chooses an action $a$ randomly from its set of actions $\mathcal A_i$ with probability $\epsilon$ (explore) that decreases after each episode and selects $a \in \arg\max_{a'\in\mathcal A_i}  Q_i(o,a'|\theta)$ with probability $1-\epsilon$ (exploit). 

\subsection{Hysteretic-MADRL}
This approach is an integration of hysteretic idea into the double DQN~\cite{MatignonLF12}. Hysteretic Q-learning is an optimistic MARL algorithm originally introduced to address maximum based learner’s vulnerability towards stochasticity by using two learning rates $\alpha$ and $\beta$, where $\beta<\alpha$ \cite{matingon07}. 
In particular, the optimistic learning idea affects the way Q values are updated. 
Given a TD error $\delta$, a hysteretic Q-value update is performed as follows:
\begin{equation}
    Q_i(o,a) \longleftarrow \begin{cases} Q_i(o,a) + \alpha\delta\ &\text{if}\ \delta>0\\
    Q_i(o,a) + \beta\delta \ &\text{otherwise},
    \end{cases}
\end{equation}
where $\beta$ is used to reduce the impact of negative Q-value updates while learning rate $\alpha$ is used for positive updates.
In \cite{PalmerTBS18}, the authors implemented a scheduled hysteretic DQN (Hysteretic-MADRL) that uses $n$ pre-computed learning rates $\beta_1,...,\beta_n$ for the double DQN, where $\beta_n$ approaches $\alpha$, and $\beta_j = d^{n-j}\beta_n$ with $d\in(0,1]$. 

\subsection{Lenient-MADRL}
In \cite{PalmerTBS18}, the authors proposed a lenient DQN (Lenient-MADRL) by incorporating lenient learning into the double DQN. The lenient learning was introduced in \cite{PotterJ94} that updates multiple agents' policies towards an optimal joint policy simultaneously by letting each agent adopt an optimistic disposition at the initial exploration phase. More precisely, lenient agents keep track of the temperature $T(o,a)$ for each observation-action pair, which is initialized by a defined maximum temperature value, and compute lenient functions by
\begin{equation} \label{eq:temperature_lenient}
    l_i(o,a) = 1 - e^{-K*T_i(o,a)},
\end{equation}
where $K$ is a constant determining how the temperature affects the decay in leniency. The temperature $T_i$ can be simply decayed by a discount factor $\theta\in[0,1]$ such that $T_i(o,a) \leftarrow \theta T_i(o,a)$. \cite{WeiL16} deployed the average temperature of the agent's next state in updating of the current temperature
\begin{equation}
    T_i(o,a) \longleftarrow \beta\begin{cases}
     T_i(o,a)\ & \text{if}\ s'\ \text{is the terminal state}\\
    (1-\nu)T_i(o,a) + \nu\bar T_i(o')\ & \text{otherwise},
    \end{cases}
\end{equation}
where $\bar T_i(o') = \frac{1}{|\mathcal A_i|}\sum_{a'\in\mathcal A_i}T_i(o',a')$. The Q-value is then updated by
\begin{equation}\label{lenient_cond}
    Q_i(o,a) \longleftarrow \begin{cases} Q_i(o,a) + \alpha\delta\ &\text{if}\ \delta>0\ \text{or}\ x > l_i(o,a)\\
    Q_i(o,a) &\text{otherwise},
    \end{cases}
\end{equation}
where the random variable $x\sim U(0,1)$ guarantees that a negative update $\delta$ is performed with a probability $1-l_i(o,a)$. Recently, the idea of the lenient Q-learning has been successfully applied to the double DQN \cite{PalmerTBS18}. In particular, \cite{PalmerTBS18} proposed 
Lenient-MADRL that minimizes the loss function \eqref{lossDQN} with samples satisfying the conditions in \eqref{lenient_cond}.

While these aforementioned DRL approaches have been successful in solving \texttt{CMOTP} tasks, it is still unclear to what extent they can perform in extended variations of stochastic reward environments, and in particular, in problems with rare events. In this work, we focus on improving the performance of the independent agents in fully cooperative tasks under various settings of stochastic rewards. Our approaches are to leverage cognitive models of human decision making and integrate MADRL concepts to help the agents enhance coordination among each other in the face of coping with diverse situations of stochastic and rare rewards.

\section{Multi-Agent IBL Models} 
IBLT is a theory of decisions from experience, developed to explain human learning in dynamic decision environments~\cite{GONZALEZ03}. IBLT provides a decision making algorithm and a set of cognitive mechanisms that can be used to implement computational models of human decision learning processes. The algorithm involves the recognition and retrieval of past experiences (i.e., instances) according to their similarity to a current decision situation, the generation of expected utility of the various decision alternatives, and a choice rule that generalizes from experience. An ``instance" in IBLT is a memory unit that results from the potential alternatives evaluated. These are memory representations consisting of three elements: a situation (a set of attributes that give a context to the decision, or observation $o$); a decision (the action taken corresponding to an alternative in state $s$, or action $a$); and a utility (expected utility or experienced outcome $x$ of the action taken in a state). 

In particular, for the agent $i$, an option $k=(o,a)$ is defined by taking action $a$ after observing state $s$.
At time $t$, assume that there are $n_{k,t}$ different generated instances $(k,x_{j,k,t})$ for $j = 1,...,n_{k,t}$, corresponding to selecting $k$ and achieving outcome $x_{j,k,t}$. Each instance $j$ in memory has an \textit{Activation} value, which represents how readily available that information is in memory, and it is determined by similarity to past situations, recency, frequency, and noise~\cite{ANDERSON14}. 
Here we consider a simplified version of the Activation equation which only captures how recently and frequently instances are activated:
 \begin{equation}\label{eq:activation}
 \begin{array}{l}
     \Lambda_{j,k, t} = \ln{\left(\sum\limits_{t' \in T_{j,k,t} }(t-t')^{-d}\right)}  + \sigma\ln{\frac{1-\xi_{j,k,t}}{\xi_{j,k,t}}},
\end{array}
 \end{equation}
where $d$ and $\sigma$ are the decay and noise parameters, respectively, and $T_{j,k,t}\subset \{0,...,t-1\}$ is the set of the previous timestamps in which the instance $j$ was observed. The rightmost term represents the Gaussian noise for capturing individual variation in activation, and $\xi_{j,k,t}$ is a random number drawn from a uniform distribution $U(0, 1)$ at each timestep and for each instance and option.

Activation of an instance $j$ is used to determine the probability of retrieval of an instance from memory.
% executing action $A$ at state $S$. 
The probability of an instance $j$ is defined by a soft-max function as follows
 \begin{equation} \label{eq:retrieval_prob}
     p_{j,k,t} = \frac{e^{\Lambda_{j,k, t}/\tau}}{\sum_{j' = 1}^{n_{k,t}}e^{\Lambda_{j',k, t}/\tau}},
 \end{equation}
where $\tau$ is the Boltzmann constant (i.e., the ``temperature") in the Boltzmann distribution.
For simplicity, $\tau$ is often defined as a function of the same $\sigma$ used in the activation equation $\tau= \sigma\sqrt{2}$.

The expected utility of option $k$ is calculated based on a mechanism called \textit{Blending}~\cite{LEBIERE99} as specified in IBLT~\cite{GONZALEZ03}, using the past experienced outcomes stored in each instance. Here we employ the Blending calculation for the agent $i$ as defined for choice tasks~\cite{LEJARRAGA12,GONZALEZ11}:
 \begin{equation} \label{eq:blended_value}
     V_{i,k,t} = \sum_{j=1}^{n_{k,t}}p_{j,k,t}x_{j,k,t}.
 \end{equation}
The choice rule is to select the option that corresponds to the maximum blended value. In particular, at the $l$-th step of an episode, the agent $i$ select the option $(o_{i,l},a_{i,l})$ with
\begin{equation}
  a_{i,l} \in \arg\max_{a\in \mathcal A_i}  V_{i,(o_{i,l},a),t}
\end{equation}

% \subsection{Multi-Agent IBL models for Markov games}
Our proposed Multi-Agent IBL (MAIBL) Models are developed to deal with fully cooperative tasks that can be described as a Markov game \cite{Shapley1095} where all agents receive the same rewards. Also, \cite{Nguyen21} proposed an IBL model (\textbf{IBL-TD}) that uses the TD-learning mechanism of RL models, to estimate the outcome of an action as follows:

\begin{equation}
    x_{i,l} \leftarrow V_{i,(o_{i,l},a_{i,l}),t} + \alpha \delta_{i,l},
\label{eq:IBLTD_update}
\end{equation}

where $\alpha$ is a learning rate and $\delta_{i,l}$ is an TD error defined by:
\begin{equation}
    \delta_{i,l} = r_{i,l+1}+\gamma \max_{a\in\mathcal A_i} V_{i,(o_{i,l+1},a),t}-V_{i,(o_{i,l},a_{i,l}),t}.
\label{eq:IBLTD_err}
\end{equation}

We refer to \cite{NGUYEN20,NGUYEN2020ICCM,nguyen2021theory} to demonstrate how to investigate IBLT for multi-state environments.

The MAIBL process is described in Algorithm \ref{alg:IBL}. We propose three MAIBL algorithms that rely on the IBL-TD, and are enhanced with a \eg exploration strategy to deal with fully cooperative tasks in MAS (Greedy-MAIBL); and a Hysteretic-MAIBL and Lenient-MAIBL, which are respectively the integration of hysteretic and lenient concepts from MADRL into the MAIBL models. 

\begin{algorithm}[!htpb]
\caption{Multi-Agent Instance-Based Learning (MAIBL) Process} 
\label{alg:IBL}
\SetKwInput{KwInput}{Input}
\DontPrintSemicolon

% \begin{algorithmic}[1] 
\KwInput{default utility $x_0$, a memory dictionary $\mathcal M= \{\}$, global counter $t = 1$, step limit $L$}
\Repeat{task stopping condition}{  
\tcp*[l]{Loop for each episode}
Initialize a counter (i.e., step) $l=0$ and each agent $i$ observes $o_{i,l}$ from state $s_l$
%   \medskip

\While{$s_l$ is not terminal and $l<L$}{

 Each agent $i$ chooses an action $a_{i,l}\in\arg\max_{a\in \mathcal A_i}V_{i,(o_{i,l},a),t}$, where $V_{i,(o_{i,l},a),t}$ is computed by Equation \eqref{eq:blended_value}\;
 Take joint action $[a_{1,l},...,a_{m,l}]$, move to state $s_{l+1}$, each agent $i$ observes $o_{i,l+1}$, and gets outcome $x_{i,l}$\;
Store timestamp $t$ to instances $(o_{i,l},a_{i,l},x_{i,l})$ for $i=1,...,m$\;
$l \leftarrow l+1$  and $t \leftarrow t+1$\;
   }
 }
% \end{algorithmic}
\end{algorithm}

\subsection{Greedy-MAIBL}
This model, called Greedy-MAIBL, integrates a decreasing $\epsilon$-greedy Boltzmann exploration strategy into the IBL-TD algorithm~\cite{Nguyen21}. More specifically, we improve the natural exploration process of IBL in IBL-TD by employing the $\epsilon$-greedy Boltzmann exploration strategy before applying the IBL's blended value. 
The motivation behind the integration is that in fully-cooperative multi-agent problems, the agents only receive feedback upon their accomplishment as a team, and no immediate feedback is available upon miscoordination or a subtask completion.
Moreover, the probability of having the best joint action (i.e. agreement in selecting actions) is low. For example, in such \texttt{CMOTP} tasks (see Section~\ref{subsec:CMOTP}), the agents only have a 20\% chance of choosing identical actions per state transition, considering that there are five possible actions: moving up, down, left, right or stay. As a result, thousands of state transitions are often required to accomplish the task and receive a reward while the agents explore the environment.
Therefore, exploration plays a vital role in cooperative MAS.

\begin{algorithm}[!htpb]
\caption{Greedy-MAIBL} 
\label{algo1}
\SetKwInput{KwInput}{Input}
\DontPrintSemicolon

% \begin{algorithmic}[1] 
\KwInput{default utility $x_0$, a memory dictionary $\mathcal M= \{\}$, global counter $t = 1$, $\epsilon$, $\eta\in (0,1)$, $\alpha$, and step limit $L \in \mathbb{N}^+$}
\Repeat{task stopping condition}{  
\tcp*[l]{Loop for each episode}
Initialize episode step counter $l=0$, $\epsilon\leftarrow \eta\epsilon$, and each agent $i$ observes $o_{i,l}$ at state $s_l$
%   \medskip

\While{$s_l$ is not terminal and $l<L$}{

 Each agent $i$ chooses $a_{i,l}\in\mathcal A_i$ randomly according to $p(o_{i,l},a_{i,l})$ with probability $\epsilon$, and $a_{i,l}\in \arg\max_{a\in\mathcal A_i}V_{i,(o_{i,l},a),t}$ with probability $1-\epsilon$, where $p(o,a) = e^{V_{i,(o,a),t}/T}/\sum_{a'\in\mathcal A_i}  e^{V_{i,(o,a'),t}/T}$\;
Take actions $[a_{1,l},...,a_{m,l}]$, move to state $s_{l+1}$, each agent $i$ observes $o_{l+1}$ at state $s_{l+1}$, and   reward $r_{l+1}$\;
Each agent $i$ computes the TD error $\delta_{i,l}$ by  Equation \eqref{eq:IBLTD_err} and estimates an outcome $x_{i,l}$  by Equation \eqref{eq:IBLTD_update}\;

Each agent $i$ stores timestamp $t$ to instance $(o_{i,l},a_{i,l},x_{i,l})$\; % to $I_k$ \;
$l \leftarrow l+1$  and $t \leftarrow t+1$\;
   }
 }
% \end{algorithmic}
\end{algorithm}

The idea of a decreasing $\epsilon$-greedy Boltzmann exploration strategy is that at time $t$, the agent $i$ chooses an action $a$ randomly according to $p(o,a)$ from its set of actions $\mathcal A_i$ with probability $\epsilon$ (explore) that decreases after each episode and selects $a \in \arg\max_{a'\in\mathcal A_i}  V_{i,(o,a'),t}$ with probability $1-\epsilon$ (exploit), where $V_{i,(o,a),t}$ is the Blended value of the agent $i$ for an option $(o,a)$ with $o$ being its observation, and $p(o,a) = e^{V_{i,(o,a),t}/T}/\sum_{a'\in\mathcal A_i}  e^{V_{i,(o,a'),t}/T}$ with $T\geq 0$ being a temperature parameter.
The full Greedy-MAIBL algorithm is described in Algorithm \ref{algo1}.

\subsection{Hysteretic-MAIBL}
The Hysteretic-MAIBL model is built upon the Greedy-MAIBL algorithm by incorporating an optimistic learning assumption of hysteretic Q-learning. 
% The incorporation of the hysteretic mechanism (i.e. two learning rates are used to control the outcome updates while relying on an optimistic form of learning) into the Greedy-MAIBL.
In the context of MARL, an intuitive way of interpreting the assumption is that an agent selects any action
it finds suitable with the expectation that the other agents also choose the best match accordingly~\cite{lauer2000algorithm}. 
Under this assumption, when playing actions, the agents prefer superior results (i.e. TD error is greater than 0), and hence the superior results are updated with a higher learning rate.
More specifically, the Hysteretic-MAIBL algorithm uses two learning rates $\alpha>\beta$ for the increase and decrease of outcomes instead of only one, as in the Greedy-MAIBL.
The Hysteretic-MAIBL algorithm for cooperative MAS is specified in Algorithm \ref{algo2}.

% \paragraph{Hysteretic-MAIBL.} Algorithm description:
\begin{algorithm}[!htpb]
\caption{Hysteretic-MAIBL} 
\label{algo2}
\SetKwInput{KwInput}{Input}
\DontPrintSemicolon

% \begin{algorithmic}[1] 
\KwInput{default utility $x_0$, a memory dictionary $\mathcal M= \{\}$, global counter $t = 1$, $\epsilon$, $\eta\in (0,1)$, $\alpha>\beta$, and step limit $L \in \mathbb{N}^+$}
\Repeat{task stopping condition}{  
\tcp*[l]{Loop for each episode}
Initialize episode step counter $l=0$, $\epsilon\leftarrow \eta\epsilon$, and each agent $i$ observes $o_{i,l}$ at state $s_l$
%   \medskip

\While{$s_l$ is not terminal and $l<L$}{

Each agent $i$ chooses $a_{i,l}\in\mathcal A_i$ randomly according to $p(o_{i,l},a_{i,l})$ with probability $\epsilon$, and $a_l\in \arg\max_{a\in\mathcal A_i}V_{i,(o_{i,l},a),t}$  with probability $1-\epsilon$, where $p(o,a) = e^{V_{i,(o,a),t}/T}/\sum_{a'\in\mathcal A_i}  e^{V_{i,(o,a'),t}/T}$\;
Take actions $[a_{1,l},...,a_{m,l}]$, move to state $s_{l+1}$, each agent $i$ observes $o_{i,l+1}$ at state $s_{l+1}$, and reward $r_{i,l+1}$\;
Each agent $i$ computes the TD error $\delta_{i,l}$ by Equation \eqref{eq:IBLTD_err}\;
Each agent $i$ estimates an outcome $x_{i,l}$ by
\begin{equation}
    x_{i,l} \longleftarrow \begin{cases} V_{i,(o_l,a_l),t} + \alpha\delta_{i,l} \ &\text{if}\ \delta_{i,l} >0\\
    V_{i,(o_l,a_l),t} + \beta\delta_{i,l} \ &\text{otherwise}
    \end{cases}
\end{equation}
\;
Store timestamp $t$ to instances $(o_{i,l},a_{i,l},x_{i,l})$ for $i=1,...,m$ \;% to $I_k$\;
$l \leftarrow l+1$  and $t \leftarrow t+1$\;
   }
 }
% \end{algorithmic}
\end{algorithm}

\subsection{Lenient-MAIBL}
The Lenient-MAIBL approach incorporates the concept of lenient learning in MARL into Greedy-MAIBL. 
The idea of leniency here is that initially, none of the agents have a good understanding of their best joint actions. Therefore they must be lenient to the foolish and arbitrary actions being made by their collaborators at the beginning~\cite{WeiL16}. 
More specifically, the leniency is affected by the frequency of visiting state-action pairs. For each state-action pair, initially, it is visited less frequently, resulting in the higher value of $T(o,a)$. The higher the value of $T(o,a)$, the more lenient the agent is (see Eq.~\eqref{eq:temperature_lenient}). That is, it ignores inferior results (i.e. the ones that result in negative TD error). 
When the state-action pair has been encountered frequently enough, more results are updated no matter what. 
The whole procedure of the Lenient-MAIBL approach is depicted in Algorithm \ref{algo3}.

% \paragraph{Lenient-MAIBL.} Algorithm description of Lenient-MAIBL:

\begin{algorithm}[!htpb]
\caption{Lenient-MAIBL} 
\label{algo3}
\SetKwInput{KwInput}{Input}
\DontPrintSemicolon

% \begin{algorithmic}[1] 
\KwInput{default utility $x_0$, a memory dictionary $\mathcal M= \{\}$, global counter $t = 1$, $\epsilon$, $\eta\in (0,1)$, $\alpha$, step limit $L \in \mathbb{N}^+$, maximum temperature value $T_{\max}$, $K$, $\theta$, and $\nu$}
\Repeat{task stopping condition}{  
\tcp*[l]{Loop for each episode}
Initialize episode step counter $l=0$, $\epsilon\leftarrow \eta\epsilon$, and each agent $i$ observes $o_{i,l}$ at state $s_l$
%   \medskip

\While{$s_l$ is not terminal and $l<L$}{

Each agent $i$ chooses $a_{i,l}\in\mathcal A_i$ randomly according to $p(o_{i,l},a_{i,l})$ with probability $\epsilon$, and $a_{i,l}\in \arg\max_{a\in\mathcal A_i}V_{i,(o_{i,l},a),t}$ with probability $1-\epsilon$, where $p(o,a) = e^{V_{i,(o,a),t}/T}/\sum_{a'\in\mathcal A_i}  e^{V_{i,(o,a'),t}/T}$\;
Take actions $[a_{1,l},...,a_{m,l}]$, move to state $s_{l+1}$, each agent $i$ observes $o_{i,l+1}$ at state $s_{l+1}$, and reward $r_{i,l+1}$\;
Each agent $i$ computes the TD error $\delta_{i,l}$ by Equation \eqref{eq:IBLTD_err}\;
$R$ $\longleftarrow$ random real value between $0$ and $1$\;
Each agent $i$ estimates an outcome $x_{i,l}$ by
\begin{equation}
    x_{i,l} \longleftarrow \begin{cases} V_{i,(o_l,a_l),t} + \alpha{\color{black}\delta_{i,l}} \ &\text{if}\ \delta_{i,l} >0\ \text{or}\ R > 1-e^{-K*T_{i,(o_l,a_l)}} \\
    V_{i,(o_l,a_l),t} \ &\text{otherwise}
    \end{cases}
\end{equation}
\;
Store timestamp $t$ to instances $(o_{i,l},a_{i,l},x_{i,l})$ for $i = 1,...,m$\; % to $I_k$\;
Update the temperature $T$
\begin{equation}
    T_{i,(o_{i,l},a_{i,l})} \longleftarrow \theta\begin{cases} (1-\nu)T_{i,(o_{i,l},a_{i,l})} + \nu \bar{T}_{i,o_{i,l+1}} \ &\text{if $s_{l+1}$ is not the terminal state} \\
    T_{i,(o_{i,l},a_{i,l})} \ &\text{otherwise},
    \end{cases}
\end{equation}
where $\bar{T}_{i,o_{i,l+1}} = \frac{1}{|\mathcal A_i|}\sum_{a\in\mathcal A_i}T_{i,(o_{i,l+1},a)}$\;
$l \leftarrow l+1$  and $t \leftarrow t+1$\;
   }
 }
% \end{algorithmic}
\end{algorithm}

% %--------------------------------------
\section{Experiments}
To make our focus concrete, we specifically consider one of the prominent examples of fully cooperative games which is the Coordinated Multi-agent Object Transportation Problems (\texttt{CMOTP}s)~\cite{Busoniu2010,PalmerTBS18,tuci2006cooperation} with the presence of \textit{stochastic} rewards. 
In such environments, we examine how well IBL-based models can learn and adapt to the other teammates' behavior to accomplish the task without communicating during the learning process.
We compare our proposed models with the three state-of-the-art algorithms in \texttt{CMOTP}s (see Section~\ref{sec:background}): Decreasing $\epsilon$-greedy double Deep Q-Network algorithm (Greedy-MADRL) \cite{HasseltGS16}, Scheduled Hysteretic Deep Q-Network algorithm (Hysteretic-MADRL) and Lenient Deep Q-Network (Lenient-MADRL) \cite{PalmerTBS18}. 

\subsection{Coordinated Multi-Agent Object Transportation Problems}\label{subsec:CMOTP}

% \subsection{Task description}
The \texttt{CMOTP} is an abstraction of a generic task involving two agents' coordinated transportation of an item. It has been used as an illustrative demonstration of a number of MARL algorithms~\cite{Busoniu2010,PalmerTBS18}.

\begin{figure}[!htpb]
	\centering
% 	\begin{tabular}{cc}
% 	\hspace{-75pt}
% 	\begin{subfigure}[]{0.3\textwidth}
%     \includegraphics[width=\textwidth]{images/cmotp1.png}
%     \caption{Simple}
% 	\end{subfigure}
% 	\hspace{70pt}
	\begin{subfigure}[]{0.32\textwidth}
    % \subfigure{
    \includegraphics[width=\textwidth]{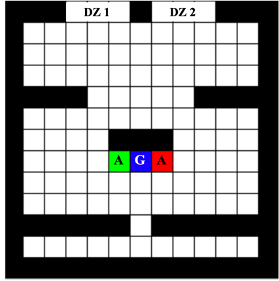} 
    \caption{Pickup}
    % \vspace{-14pt}
    % }
	\end{subfigure}
	\hspace{10pt}
 \begin{subfigure}[]{0.32\textwidth}
    % \subfigure{
    \includegraphics[width=\textwidth]{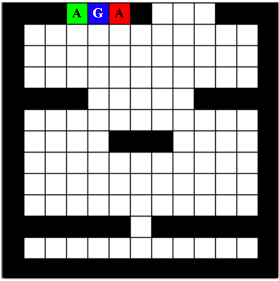} 
    \caption{Delivery}
    
    % }
	\end{subfigure}
%     % \hspace{70pt}
% 	\begin{subfigure}[]{0.42\textwidth}	
%     % \subfigure{
    \includegraphics[width=0.4\textwidth]{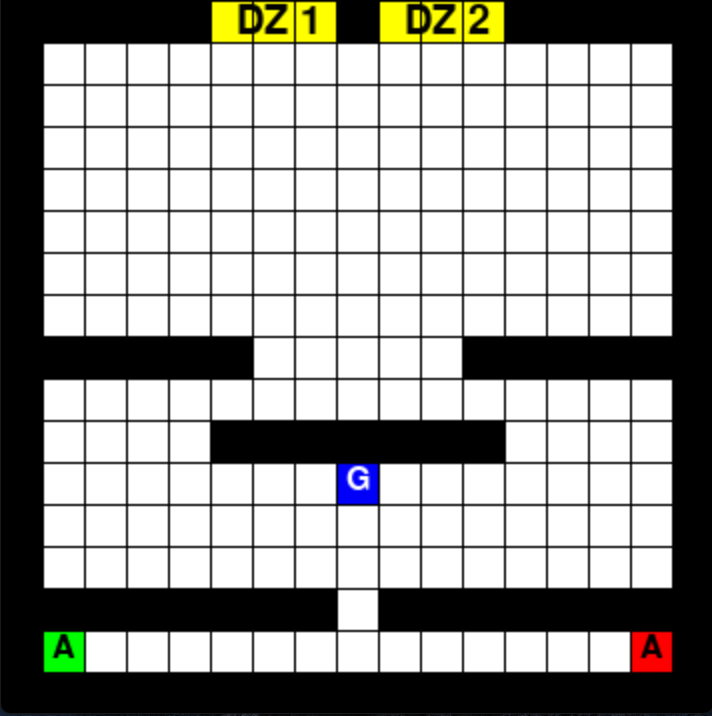}
    % \caption{stochastic}
    % }
% 	\end{subfigure}
% 	\end{tabular}
	\caption{Coordinated Multi-Agent Object Transportation Problem. }
	
	\label{cmotp}
\end{figure}

In particular, the \texttt{CMOTP} is simulated in a gridworld, that is, in a two-dimensional discrete grid with $16\times16$ cells as illustrated in Fig~\ref{cmotp}. The idea of the task is that the agents (represented by letter A) have to navigate and transport a target item (G) to one of two drop-zone(s) (yellow areas) while avoiding obstacles (represented by black cells). In other words, the agents share a common interest in delivering an item (G) to the drop zone. Thereby, the agents must coordinate themselves to get an equally shared reward; otherwise, they fail and get a zero reward.

To complete the tasks, the agents must exit the room individually to locate and collect item G. Pickup is only possible when the two agents stand on item G's left- and right-hand sides in the grid (Fig.~\ref{cmotp} a). Once the two agents have grasped either side of the item, they can move it. 
The task is fully cooperative, as the item can only be transported when both agents successfully grab the item by always being side-by-side and deciding to move in the same direction. The agents choose to \textit{stay in place}, \textit{move left}, \textit{right}, \textit{up}, or \textit{down}, and can move only one cell at a time.
Agents can only move to an empty cell, and if both try to move to the same cell, neither moves.

Agents only receive a positive reward after placing the item inside the dropzone (illustrated in Fig.~\ref{cmotp} b). In case there are multiple drop zones, the agents' goal is to drop the item into the drop zone yielding the highest expected reward. To encourage agents to complete the task as quickly as possible, agents are penalized for walking into an obstacle ($-0.05$) and deciding to stand still ($-0.01$). 

\subsection{Experimental Design}
\paragraph{Stochastic Reward Scenarios}
We characterize four different stochastic reward scenarios inspired by the study of decisions from experience with rare events in risky choice~\cite{hertwig2004decisions}. That is, the scenarios are selected to represent a diverse variety of situations with stochastic rewards in order to understand better how the agents not only coordinate to accomplish the task but also learn to deal with diverse situations of stochastic rewards. The characteristics of the scenarios considered are determined by whether the optimal option is deterministic (safe) or stochastic (risky) and the probability of getting the high value of the stochastic option. These scenarios are summarized in Table \ref{tab:reward}.

\begin{table}[!htpb]
\label{scenearios}
%$V3(1), V8(2), V6(3) V4(4)$
% \begin{sidewaystable}
% \begin{adjustbox}{angle=90}
    \centering
    \resizebox{1\textwidth}{!}{%
    \begin{tabular}{c|ll|ll} 
\toprule 
\multirow{ 2}{*}{Scenarios} & \multicolumn{2}{|c|}{Zones} & \multicolumn{2}{|c}{Expected reward} \\
\cline{2-5}
 & High (DZ1) & Low (DZ2) & High (DZ1) & Low (DZ2) \\ 
 \midrule 
 1 & {\color{black} $\mathbb P(r_1 = 0.8) = 1$}& {\color{black} $\mathbb P(r_2 = 1) =0.6$ and $\mathbb P(r_2 = 0.4) = 0.4$} & 0.8 & 0.76 \\
2 & {\color{black} $\mathbb P(r_1 = 0.8) = 1$} & {\color{black} $\mathbb P(r_2 = 7) = 0.1$ and $\mathbb P(r_2=0.06)=0.9$}& 0.8 & 0.754 \\
 3 & {\color{black} $\mathbb P(r_1 = 4) = 0.8$ and $\mathbb P(r_1 = 0) = 0.2$} & {\color{black} $\mathbb P(r_2=3)=1$} & 3.2 & 3\\
 4 & {\color{black} $\mathbb P(r_1 = 32) = 0.1$ and $\mathbb P(r_1 = 0) = 0.9$} & {\color{black} $\mathbb P(r_2=3)=1$} & 3.2 & 3 \\

 \bottomrule
\end{tabular}
}
    \caption{Stochastic scenarios.}
    \label{tab:reward}
% \end{adjustbox}
% \end{sidewaystable}
% }
\end{table} 

\begin{itemize}
    \item \textbf{Scenarios 1}: DZ1 is a deterministic zone always giving a reward $0.8$, whereas the DZ2 is a stochastic zone returning a reward of 1 on 60\% of occasions and 0.4 on the other 40\%. Therefore, the optimal joint policy is that the agents deliver the item to the deterministic DZ1 yielding a reward of 0.8, as opposed to an average reward of 0.76 for DZ2.
    \item \textbf{Scenarios 2}: DZ1 is a deterministic zone giving a reward of 0.8, while the DZ2 is the stochastic zone that returns a higher reward of 7 on a low probability of 0.1 and 0.06 on the other 0.9. The optimal joint policy is the deterministic DZ1 yielding a reward of 0.8, as opposed to DZ2 yielding an average reward of 0.754.
    \item \textbf{Scenarios 3}: DZ1 is a stochastic zone giving a reward of 4 on 80\% of occasions and 0 otherwise, whereas the DZ2 returns a reward of 3. The optimal joint policy is the stochastic DZ1 yielding an expected reward of 3.2, as opposed to DZ2, that only returns a reward of 3.
    \item \textbf{Scenarios 4}: DZ1 is stochastic, giving a reward of 32 on a low probability of $0.1$ and 0 otherwise, and the DZ2 is a deterministic zone giving a reward of 3. The optimal joint policy is the stochastic DZ1 yielding an expected reward of 3.2, as opposed to DZ2, which only has a reward of 3.
\end{itemize}

\subsection{Measures} For each of the models considered in the experiment, we measured their behavior and performance with respect to the following metrics:

(1) \textit{Average Proportion of Maximization-PMax}: the average proportion of episodes in which the agents delivered the item to the optimal zone, that is,
\begin{equation}
    PMax = \frac{1}{\#run}\sum_{i=1}^{\#run}\frac{\#episode_o^i}{\#episode},
\end{equation}
where $\#run$ is the number of runs, $\#episode_o^i$, and $\#episode$ are respectively the number of episodes of the $i$-th run that the agents delivered the item to the optimal zone (called \textit{optimal} episodes) and the total number of episodes. {\color{black}This metric essentially captures the effectiveness of agents as a team by delivering the item to the optimal zone.}

(2) \textit{Average Coordination Rate-PCoordinate}: the average proportion of  steps the agents successfully move together (i.e. they both move in the same direction) to the total number of steps after they stick together with the item from the pickup point, multiplied by Pmax, namely {\color{black}
\begin{equation}
    PCoordinate = \frac{1}{\#run}\sum_{i=1}^{\#run}\frac{1}{\#episode}\sum_{e=1}^{\#episode_o^i}\frac{\#step_m^{i,e}}{\#step_s^{i,e}},
\end{equation} }
where $\#step_m^{i,e}$ is the number of steps that the agents successfully move the item, $\#step_s^{i,e}$ is the total number of steps after they stick together with the item from the pickup point at {\color{black}the optimal episode} $e$. {\color{black}This metric represents how well the agents coordinate with each other-- that is, how many times they reach a consensus on selecting their movement direction, throughout the process of dropping the item into the optimal zone.}

(3) \textit{Average Discounted Reward-Efficiency}: the discounted reward is defined by $\gamma^lr$ in which a positive reward $r$ is discounted by a discount factor $\gamma$ increased to the power of the number of steps taken $l$, multiplied by PMax, namely {\color{black}
\begin{equation} 
    Efficiency = \frac{1}{\#run}\sum_{i=1}^{\#run}\frac{1}{\#episode}\sum_{e=1}^{\#episode_o^i}\frac{\gamma^{\#step^{i,e}}r^{i,e}}{R},
\end{equation}  }
where $\#step^{i,e}$ and $r^{i,e}$ are respectively the total number of steps taken by the two agents in a team and the collective reward at episode $e$ of the run $i$, and $R$ is the high expected reward. {\color{black}This metric captures the efficiency of agents as a team in delivering the item to the optimal zone. Indeed, the metric considers not only the rewards obtained (i.e. how effective the agents are) but also how many steps are taken to get the reward (i.e. how quickly the agents learn to successfully accomplish the task).}

(4) \textit{Number of Steps-Step}: the average total number of steps taken by the two agents in a team, namely
\begin{equation}
    Step = \frac{1}{\#run}\sum_{i=1}^{\#run}\frac{1}{\#episode_o^i}\sum_{e=1}^{\#episode_o^i}\#step^{i,e};
\end{equation}
{\color{black} This metric evaluates the total number of steps taken by the agents to successfully drop the item into the optimal zone. In particular, it counts the steps when the agents are successful in moving in the same direction as well as when they are not.}

(5) \textit{Maximum Pickup Steps-MStep}: the average maximum number of steps taken by both agents to locate and pick up the item, namely
\begin{equation}
    MStep = \frac{1}{\#run}\sum_{i=1}^{\#run}\frac{1}{\#episode_o^i}\sum_{e=1}^{\#episode_o^i}\max(\#step^{i,e}_1,\#step^{i,e}_2),
\end{equation}
where $\#step^{i,e}_1$ and $\#step^{i,e}_2$ are respectively the numbers of steps of the run $i$ taken by the two agents to pick up the item at optimal episode $e$.
{\color{black} This metric examines the maximum number of  steps that one could take to find the item. 
}

(6) \textit{Difference Pickup Step-DStep}: the average difference in the number of steps taken between the two agents, taken to pick up the item, namely
\begin{equation}
    DStep = \frac{1}{\#run}\sum_{i=1}^{\#run}\frac{1}{\#episode_o^i}\sum_{e=1}^{\#episode_o^i}|\#step^{i,e}_1-\#step^{i,e}_2|.
\end{equation}
{\color{black} It is worth noting that in this measure, we only consider the episodes of the $i$-th run that the agents successfully delivered the item to the optimal zone ($\#episode_o^i$, called \textit{optimal} episodes). This metric relates to the functional delay metric~\cite{hoffman2019evaluating}, in which their teammate incurs the delay experienced by the agent after locating the item. 
The higher value, the longer it takes for an agent to wait for another agent to get to the pickup place.}

\subsection{Model Parameters}
We ran experiments using default parameter values for each model. That is, these values are commonly used in the literature. 
For the IBL part and the decreasing $\epsilon$-greedy strategy of the three IBL-based models, we used decay $d=0.5$, noise $\sigma=0.25$, default\_utility = 0.1, initial epsilon $\epsilon = 1$, decreasing factor $\eta = 0.999$, T = 0.8, discount factor $\gamma = 0.99$, and learning rate $\alpha = 0.5$. For Hysteretic-IBL, we need an additional learning $\beta = 0.01$ while Lenient-IBL requires four more parameters $T_{\max} = 2, K = 1, \theta = 0.995$, and $\nu =0.1$. 
Importantly, none of the parameters in our models were optimized, whereas for those in the comparative Hysteretic-MADRL and Lenient-MADRL algorithms, including hyper-parameters we used the same values as suggested in the previous work~\cite{PalmerTBS18}. 
We do not report the parameter values of these models here, please see Table 1 in the paper of~\cite{PalmerTBS18} for more details.

We conducted 30 runs of $1000$ episodes per run. An episode terminates when a $5000$-step limit is reached or when the agents successfully place the item inside the drop zone.  

%%%----------------------------------------
\section{Results}
\subsection{Overall performance of MAIBL and MADRL models}

Table~\ref{table:performance} reports the aggregate performance metrics averaged over all episodes, for each of the four CMOTP scenarios. These results clearly indicate that the MAIBL models perform better than the MADRL models in all scenarios.

\begin{table}[!htpb]
    \centering
\resizebox{0.99\textwidth}{!}{%
    \begin{tabular}{c|l|rrr|rrr} 
\toprule 
\multirow{ 2}{*}{Scenario} & \multirow{ 2}{*}{Metric} & \multicolumn{3}{|c|}{MAIBL} & \multicolumn{3}{|c}{MADRL} \\
\cline{3-8}
 & & Greedy & Hysteretic & Lenient & Greedy & Hysteretic & Lenient \\ 
 \midrule
 \multirow{2}{*}{1} & 
 % PMax & \textbf{0.801} & 0.163 & 0.294 & 0.210 & 0.099 & 0.332 \\
 % & & (0.099) & (0.219) & (0.135) & (0.127) & (0.031) & (0.112) \\
 PMax & \textbf{0.801} (0.099) & 0.163 (0.219) & 0.294 (0.135) & 0.210 (0.127) & 0.099 (0.031) & 0.332 (0.112)\\
 
% &  Efficiency & \textbf{0.195} & 0.006 & 0.037 & 0.002 & 0.002 & 0.070 \\
% & & (0.026) & (0.009) & (0.014) & (0.002) & (0.001) & (0.019) \\
&  Efficiency & \textbf{0.195} (0.026) & 0.006 (0.009) & 0.037 (0.014) & 0.002 (0.002) & 0.002 (0.001) & 0.070 (0.019)  \\
\vspace{0.1cm}
 0.8, 1 & PCoordinate & \textbf{0.306} (0.009) & 0.039 (0.005) & 0.087 (0.008) & 0.046 (0.002) & 0.022 (0.001) & 0.107 (0.011) \\
 % & & (0.009) & (0.005) & (0.008) & (0.002) & (0.001) & (0.011) \\
 1, 0.6/0.4, 0.4 & Step & \textbf{322.4} (49.9) & 1578.3 (555.2) & 659.4 (175.1) & 1823.1 (417.1) & 1514.1 (235.3) & 603.7 (176.2)\\
 % & & (49.9) & (555.2) & (175.1) & (417.1) & (235.3) & (176.2) \\
&			MStep & \textbf{121.6} (34.7) & 514.2 (155.8) & 194.2 (44.8) & 809.2 (120.9) & 622.4 (104.8) & 282.1 (84.1)\\
% & & (34.7) & (155.8) & (44.8) & (120.9) & (104.8) & (84.1) \\
&			DStep & \textbf{30.5} (2.8) & 130.2 (47.1) & 44.1 (9.7)& 196.2 (22.9) & 144.6 (17.6) & 115.3 (36.0) \\
% &  & (2.8) & (47.1) & (9.7) & (22.9) & (17.6) & (36.0) \\
			
\midrule

\multirow{2}{*}{2} & 

PMax & \textbf{0.936} (0.025) & 0.308 (0.345) & 0.840 (0.099) & 0.806 (0.009) & 0.815 (0.018) & 0.735 (0.047)\\
% & & (0.025) & (0.345) & (0.099) & (0.009) & (0.018) & (0.047)\\
& Efficiency & \textbf{0.350} (0.027) & 0.093 (0.086) & 0.263 (0.099) & 0.169 (0.001) & 0.170 (0.001) & 0.156 (0.003) \\
% & & (0.027) & (0.086) & (0.099) & (0.001) & (0.001) & (0.003) \\
\vspace{0.1cm}
0.8, 1 & PCoordinate & \textbf{0.350} (0.019) & 0.093 (0.119) & 0.263 (0.073) & 0.169 (0.001) & 0.170 (0.003) & 0.156 (0.008) \\
% & & (0.019) & (0.119) & (0.073) & (0.001) & (0.003) & (0.008) \\
7, 0.1/0.06, 0.9 &			Step & \textbf{371.4} (42.5) & 1649.2 (945.5) & 679.8 (300.4) & 1368.7 (77.5) & 1221.8 (58.5) & 1091.4 (95.3)\\
% & & (42.5) & (945.5) & (300.4) & (77.5) & (58.5) & (95.3) \\
&			MStep & \textbf{122.1} (25.9) & 500.8 (262.5) & 252.2 (190.3) & 631.6 (43.6) & 440.9 (33.1) & 439.1 (37.5)\\
% & & (25.9) & (262.5) & (190.3) & (43.6) & (33.1) & (37.5) \\
&			DStep & \textbf{33.3} (1.3) & 125.5 (63.8) & 35.7 (2.3)& 180.1 (11.0) & 156.6 (13.1)& 149.2 (18.6) \\
% & & (1.3) & (63.8) & (2.3) & (11.0) & (13.1) & (18.6) \\
\midrule

\multirow{2}{*}{3} & PMax & 0.475 (0.375) & \textbf{0.642} (0.328) & 0.565 (0.407) & 0.166 (0.028) & 0.240 (0.021) & 0.237 (0.016) \\
% & & (0.375) & (0.328) & (0.407) & (0.028) & (0.021) & (0.016) \\
&			Efficiency & 0.134 (0.114) & \textbf{0.245} (0.125) & 0.224 (0.163) & 0.001 (0.001) & 0.004 (0.001) & 0.004 (0.002) \\
% & & (0.114) & (0.125) & (0.163) & (0.001) & (0.001) & (0.002) \\
\vspace{0.1cm}
4, 0.8/0, 0.2 &	PCoordinate & 0.224 (0.177) & \textbf{0.354} (0.177) & 0.326 (0.235) & 0.036 (0.006)& 0.052 (0.005) & 0.051 (0.004)\\
% & & (0.177) & (0.177) & (0.235) & (0.006) & (0.005) & (0.004) \\
3, 1 &			Step & 872.1 (886.5) & \textbf{696.1} (920.6) & 1078.4 (1241.2)& 1503.1 (174.9)& 1058.7 (81.7)  & 1091.4 (117.2)\\
% & & (886.5) & (920.6) & (1241.2) & (174.9) & (81.7) & (117.2) \\
&			MStep & 229.1 (167.2) & \textbf{203.8} (191.8) & 257.1 (264.7) & 637.3 (75.3) & 408.9 (43.1)& 414.7 (55.7) \\
% & & (167.2) & (191.8) & (264.7) & (75.3) & (43.1) & (55.7) \\
&			DStep & \textbf{55.2} (40.1) & 63.4 (57.5) & 71.2 (63.4) & 180.1 (17.9) & 140.2 (14.8)& 144.4 (21.3) \\
% & & (40.1) & (57.5) & (63.4) & (17.9) & (14.8) & (21.3) \\
\midrule

\multirow{2}{*}{4} & PMax & 0.016 (0.014 & \textbf{0.294} (0.394)& 0.040 (0.059) & 0.041 (0.008) & 0.021 (0.006) & 0.021 (0.004)\\
% & & (0.014) & (0.394) & (0.059) & (0.008) & (0.006) & (0.004) \\
& Efficiency & 0.000 (0.000) & \textbf{0.179} (0.247) & 0.011 (0.032)& 0.000 (0.000)& 0.000 (0.000)& 0.000 (0.000)\\	
% & & (0.000) & (0.247) & (0.032) & (0.000) & (0.000) & (0.000) \\
\vspace{0.1cm}
32, 0.1/0, 0.9 & PCoordinate & 0.004 (0.003) & \textbf{0.263} (0.360) & 0.019 (0.035) & 0.008 (0.002) & 0.004 (0.001) & 0.004 (0.001)\\
% & & (0.003) & (0.360) & (0.035) & (0.002) & (0.001) & (0.001) \\
3, 1 &			Step & 2793.8 (394.8)& \textbf{2106.3} (1500.8)& 2533.8 (1009.8)& 2195.6 (203.6)& 2847.4 (288.1)& 2674.0 (239.1)\\
% & & (394.8) & (1500.8) & (1009.8) & (203.6) & (288.1) & (239.1) \\
&			MStep & \textbf{534.7} (184.8) & 535.7 (414.6) & 603.8 (315.8)& 672.6 (117.3)& 784.4 (195.9)& 801.7 (149.9)\\
% & & (184.8) & (414.6) & (315.8) & (117.3) & (195.9) & (149.9) \\
 &			DStep & \textbf{121.2} (29.7)& 157.0 (122.7) & 160.2 (106.9)& 190.7 (27.3)& 216.1 (47.4)& 221.6 (77.1)\\
 % & & (29.7) & (122.7) & (106.9) & (27.3) & (47.4) & (77.1)\\
 \bottomrule
\end{tabular}
}
    \caption{{\color{black} Performance of the agents reported in the form of the mean value (standard deviation) with respect to the different metrics for each of the four \texttt{CMOTP} scenario}. Bold values indicate the best results.}
    \label{table:performance}
\end{table}

Overall, we observe that when the highest expected reward is associated with the deterministic zone, as in scenarios 1 and 2, the Greedy-MAIBL agents are the best performers, followed by Lenient models with regard to all the metrics. We also see that all the models perform much better in scenario 1 compared to scenario 2. This can be explained by the fact that when the stochastic zone is unlikely to return the high reward (i.e. the probability of the high value in the stochastic zone is low), it is easier for the agents to decide to select the deterministic zone with the higher expected value.
By contrast, there is more tension in choosing between the optimal zone and the stochastic zone when the high reward in the stochastic zone is more likely to happen (i.e., scenario 1). As a result, the performance of all the models was lower. 

Interestingly, in scenarios 3 and 4, wherein the stochastic zone yields the highest expected value, the Hysteretic-MAIBL turns out to achieve the best performance in terms of PMax, coordination rate, and efficiency (Discounted Reward). That said, we notice that the Greedy-MAIBL model is still more effective than the Hysteretic-MAIBL in terms of coordinating with each other to pick up the item. Compared to scenario 3, it is clear that it is more difficult for the agents in scenario 4 to select the highest expected reward zone as the high value of this zone happens rarely. 

These results show that the characteristics of stochastic reward did impact the behavior and robustness of the models, suggesting that the strength and shortcomings of each model depend on each scenario. That is, the plain Greedy-MAIBL can complete the task more successfully in the settings wherein the highest expected value belongs to the deterministic zone. However, when the highest expected value is associated with the stochastic zone, incorporating the hysteretic mechanism into Greedy-MAIBL, i.e. Hysteretic-MAIBL, becomes more effective. 

\subsection{Model's Effectiveness in Learning Optimal Delivery}
\begin{figure}[!htpb]
\begin{center}
\includegraphics[width=0.88\textwidth]{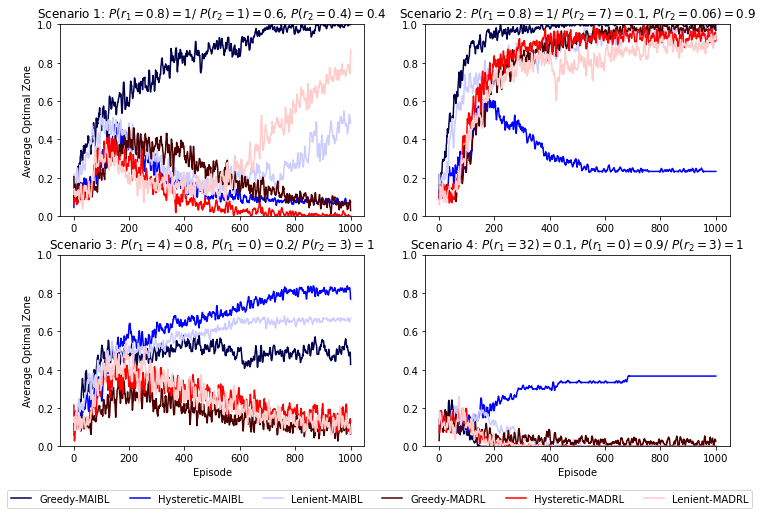}
\caption{Average proportion of maximization (PMax) over time for different agents}
\label{fig:optimal} 
\end{center}
% \vspace{-0.2in}
\end{figure}

Figure~\ref{fig:optimal} shows the performance of each model with respect to the effectiveness calculated by the optimal policy rate for 1000 episodes in each of the four scenarios. 
First, we observe that in the first two scenarios (Scenario 1 and 2), wherein the high expected reward (optimal zone) is associated with the deterministic zone, Greedy-MAIBL agents not only outperform the other models but also learn faster. Additionally, we notice that the distinction between Greedy-MAIBL and the other models is more clear in Scenario 1 than in Scenario 2. In scenario 2, all models except for Hysteretic-MAIBL compare to the Greedy-MAIBL model. Again, the explanation for this observation is that scenario 2 is a much easier decision making problem than scenario 1. The very low and common reward (0.06 with probability 0.9) of the risky option in Scenario 2, makes the discrimination between the deterministic zone much easier for most models.

Also, we see that Hysteretic-MAIBL learns faster and better in scenarios 3 and 4. These are the scenarios in which the highest expected reward is in the stochastic zone. In scenario 3, with the high probability corresponding to the higher outcome (i.e., high frequency of the high outcome), all the MAIBL models do better than the MADRL models; in contrast to scenario 4, in which the agents are misled by the high frequency of the low outcome, resulting in a decline in the performance of most models but the Hysteretic-MAIBL. 

\subsection{Models' Efficiency}

\begin{figure}[!htpb]
\begin{center}
\includegraphics[width=0.88\textwidth]{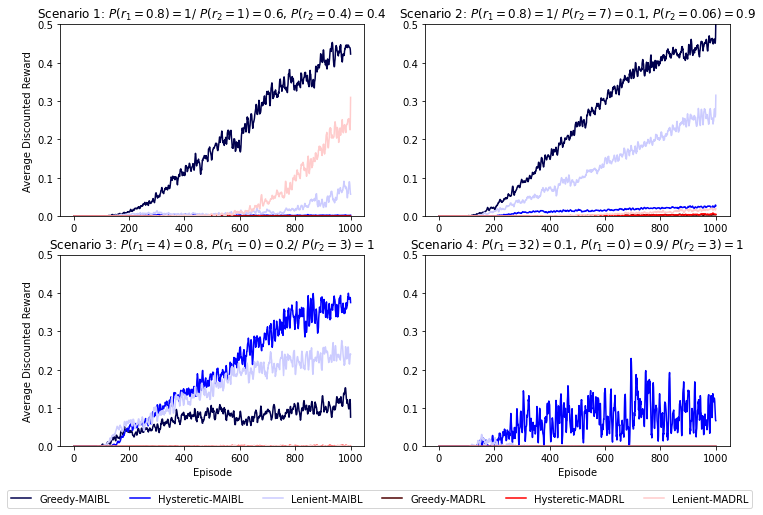}
\caption{Average discounted reward (Efficiency)}
\label{fig:reward} 
\end{center}
% \vspace{-0.2in}
\end{figure}

Fig.~\ref{fig:reward} illustrates the behavior of the models in terms of efficiency captured by the average discounted reward over 1000 episodes. We can see that after 200 episodes, Greedy-MAIBL model not only exhibits its ability to accomplish the task successfully but also is able to learn to do that with fewer steps. This pattern holds true in the first two scenarios. Interestingly, the distinction between Lenient-MAIBL and Lenient-MADRL in terms of PMax in scenarios 1 and 2 is negligible, it becomes distinct in light of average efficiency. That is, Lenient-MADRL is more efficient than Lenient-MAIBL in scenario 1. Nevertheless, it is not the case in scenario 2, wherein Lenient-MAIBL is the second-best efficient model.

In scenario 3, Hysteretic-MAIBL and Lenient-MAIBL clearly demonstrate the trend of increasing the average discounted reward over time, followed by Greedy-MAIBL. In scenario 4, by contrast, the learning curve of Hysteretic-MAIBL is the most efficient, yet it produces a large variance in the results.  

\subsection{Models' Coordination Ability}

\begin{figure}[!htpb]
\begin{center}
\includegraphics[width=0.85\textwidth]{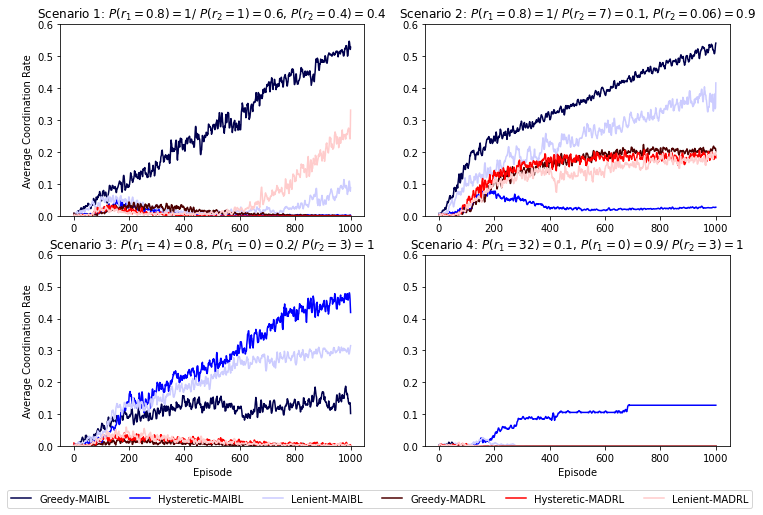}
\caption{Average coordination rate (PCoordinate)}
\label{fig:coordinate} 
\end{center}
% \vspace{-0.2in}
\end{figure}

Fig.~\ref{fig:coordinate} further demonstrates the models' coordination ability over the episodes, the average proportion of steps that the agents successfully move together, of each model across 1000 episodes. In agreement with our previous observations, Greedy-MAIBL agents have the highest coordination rates over the other agents in scenarios 1 and 2. The results also show that it is easier for the agents to coordinate in scenario 2 compared to scenario 1. That is, in scenario 1, we observe that the coordination performance of Hysteretic-MADRL and Lenient-MAIBL agents picks up after 600 episodes. In contrast, in scenario 2, it only took them about 200 episodes to see their improvement in coordination.

In scenarios 3 and 4, Hysteretic-MAIBL agents coordinate best to accomplish the task. Furthermore, all the models are extremely poor in scenario 4, wherein the optimal option is stochastic, yet the probability of getting its high value is really low. The only model that is able to handle such a challenging condition is Hysteretic-MAIBL. 

\begin{figure}[!htpb]
\begin{center}
\includegraphics[width=0.85\textwidth]{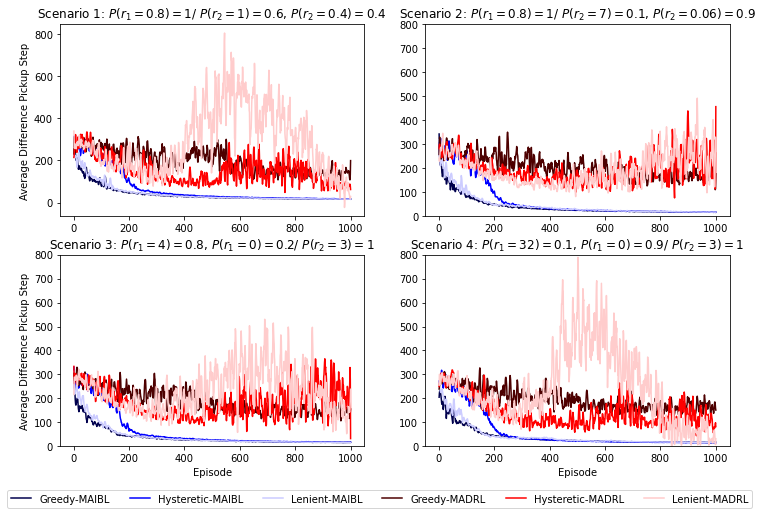}
\caption{Average difference pickup steps (DStep)}
\label{fig:dstep} 
\end{center}
% \vspace{-0.2in}
\end{figure}

\subsection{Models' Functional Delay}

{\color{black} We additionally examine the coordination ability of the models from the perspective of functional delay. In particular, this measure captures how long one agent must wait for the other agent coming to pick up the object. Simply put, the delay experienced by the agent is incurred by their teammate. 

Fig.~\ref{fig:dstep} shows the average difference in the number of steps taken between the two agents to pick up the item ($DStep$). Notably, we only calculated this measure in the episodes in which the agents accomplish the task by delivering the item to the optimal zone. A lower value of $DStep$ translates into better collaboration, as it indicates an efficient use of team members' time (steps) and a sense that their activities are smooth.}

{\color{black} The $DStep$ of the Greedy-, Lenient-, and Hysteretic-MAIBL models show a decreasing trend after 200 episodes, irrespective of scenarios. This trend can be explained by the fact that picking up the item is only a subtask of the CMOTP, and it is not directly influenced by stochastic rewards in different scenarios. Moreover, the $DStep$ of MADRL models is higher than that of MAIBL models, indicating that when MADRL agents collaborate to collect the item, the delay in the transport of the item is longer than the MAIBL models. Additionally, the results suggest that the MADRL agents fail to converge to optimal actions within 1000 episodes. In particular, the Lenient-MADRL agents show the largest disparity in the number of steps between the agents, which can be attributed to the leniency characteristics of the agents that enable them to tolerate miscoordination while the exploration is high.
}

\section{Conclusions}

Many practical real-world applications require coordination in multi-agent systems to accomplish a common goal in the absence of explicit communication. Coordination in MAS becomes particularly complicated in the presence of reward stochasticity and rare events, since miscoordination may arise when independent learners have difficulty differentiating between the teammate's exploratory behavior and the stochasticity of the environment. As a result, the current state-of-the-art MADRL models show that sub-optimal solutions emerge in non-stationary environments, due to these dynamics of coexisting agents and stochastic rewards.

This research proposes and demonstrates a solution to this problem in the current MADRL. Our solution is inspired by the human ability to adapt quickly to non-stationary environments, by the benefits of cognitive modeling approaches that have been demonstrated to capture this human behavior, and by the efficiency of RL computational concepts, such as the ``temporal difference'' adjustments that can be combined with cognitive approaches~\cite{Nguyen21}. Building on such concepts, we proposed three novel models to study cooperation and coordination behavior in MAS in the presence of stochastic rewards. The models called Greedy-MAIBL, Hysteretic-MAIBL and Lenient-MAIBL, combine the cognitive principles of the Instance-Based Learning Theory \cite{GONZALEZ03,Nguyen2022} and RL techniques to address coordination of MAS with stochastic rewards. In particular, the Greedy-MAIBL model is the enhancement of the IBL natural exploration process with the decreasing $\epsilon$-greedy Boltzmann exploration strategy due to the characteristics of the cooperative multi-agent tasks that typically require the agents to explore the environment extensively. The Hysteretic-MAIBL and Lenient-MAIBL are the methods that involve the integration of optimistic learning and leniency ideas from RL into the Greedy-MAIBL model. 

We demonstrate the merits of combining cognitive IBL and RL approaches in fully-cooperative multi-agent problems that exhibit challenging characteristics of stochastic rewards. In particular, a simulation experiment demonstrates different sources of stochasticity challenges for the MADRL models and the benefits of using MAIBL models in a Coordinated Multi-agent Object Transportation Problem. The results demonstrate these benefits on metrics including efficiency and coordination.

Our findings reveal that our proposed approaches, which are a combination of cognitive IBL and RL concepts, outperform the three state-of-the-art Deep Reinforcement Learning (DRL) algorithms in all the scenarios of stochastic rewards. These results can be attributed to the benefits of leveraging the cognitive, memory retrieval process in IBL models when applied to multi-agent problems. Indeed, the results emphasize the importance of how MAIBL models characterize {\color{black}the cognitive} frequency and recency information in the presence of stochastic reward in MAS. Although Lenient MADRL models have been advanced by incorporating the frequency information to determine how lenient an agent is supposed to be regarding others' actions, our experimental results show that it does not characterize the frequency as effectively as MAIBL models do. In MAIBL models, such frequency and recency characteristics are well captured and represented due to the declarative knowledge offered by a well-known cognitive architecture, ACT-R \cite{ANDERSON14}, which derives from well-validated human memory retrieval processes.  Additionally, it is clear that the MAIBL models also inherit the advantage of RL concepts, that is, decreasing $\epsilon$-greedy for exploration and optimistic learning of hysteretic. Thus, it is this combination of the cognitive concepts or IBL models and the computational advantages of Deep RL mechanisms that can make the MAIBL models advantageous over the MADRL in stochastic situations. 

These results can inform the selection of models that are more appropriate in specific stochastic settings and identify the characteristics of a task given an unknown reward scheme.
More concretely, the results suggest that using the simple Greedy-MAIBL model itself, the one that has the least number of parameters, it is able to surpass the other sophisticated models in the scenarios where the highest expected reward is associated with the \textit{deterministic} option regardless of the probability of returning the high value of the stochastic alternative. Our findings also indicate that Hysteretic and Leninent-based models are sensitive to the choice of the parameters and require a longer process to be able to accomplish the task. Given that hyper-parameter tuning is one of the challenging and crucial steps in the successful application of Deep RL algorithms, this work demonstrates the great benefit of using a simple Greedy-MAIBL model in settings in which the highest expected reward is associated with the deterministic alternative. 

We also learned that when the stochastic option yields the higher expected reward, incorporating Hysteretic mechanism into the Greedy-MAIBL is beneficial, leading to the fact that Hysteretic-MAIBL outperforms other models in these cases. The advantages of the Hysteretic-MAIBL model are gained from the optimistic learning idea characterized in the hysteretic model and from the frequency, and recency biases inherited in the Greedy-MAIBL model. The results suggest that in the scenarios where the stochastic alternative yields the higher expected reward, it is important for a model to integrate optimistic learning, frequency, and recency biases to effectively address fully cooperative MAS. Interestingly, the results further demonstrate that not all combinations of the IBL model and RL concepts are advantageous. Specifically, we observed that Leniency-MAIBL is not as effective as expected, suggesting that incorporating both methods of characterization of frequency in Lenient-MAIBL might not be beneficial. 

{\color{black} Arguably, one of the main goals of AI is to generate agents that can collaborate with humans and augment people's capabilities. Due to human suboptimality, prior research in collaborative scenarios has shown that agents trained to play well with other AI agents perform much worse when paired with humans~\cite{carroll2019utility}. By incorporating the cognitive characteristics of humans' decision behavior, we expect that the MAIBL models will enhance human-AI collaboration. That is, they would learn to be more adaptive to human behavior. Therefore, our future research will evaluate the performance of the MAIBL models as teammates collaborating with human participants. We further intend to experiment with heterogeneous teams wherein we team MAIBL models with different types of MADRL models.}
In future work, we also plan to investigate the robustness of our proposed models in different settings of multi-agent tasks, such as the sequentially coordinated delivery with the presence of multiple roles and expiration time. In such a problem, there are two roles of agents, and the accomplishment of the task requires the sequential collaborations of two sub-tasks.
% The first type of agent called a delivery is task with picking up a material at the storage area, delivering it, and placing it in a location in the installation area. The second type of agent called an execution agent has the role of moving to the location where the material has been placed and executing the finishing task using the delivered material as soon as possible. Delivered materials have an expiration time and become unavailable after that time. Hence, an execution agent has to execute the finishing task before the expiration of material after it has been placed; otherwise, the effort of the delivery agents is wasted. 
% Human behavior
Moreover, progress in cognitive science suggests that computational models that accurately represent human behavior would be able to collaborate with humans more successfully than models that are focused on engineering rather than the cognitive aspects of learning \cite{lake2017building}.
Given the demonstrated ability of IBL models to learn quickly and account for human learning in a wide range of tasks~\cite{Nguyen2022,nguyen2021theory}, our proposed models that are based on the combination of IBL and DRL models are expected to be an effective human partner in cooperative human-machine teaming tasks.

\section*{Reproducibility}
The code of the MAIBL models is implemented using the SpeedyIBL library \cite{Nguyen2022}. All the code for MAIBL models, simulation data, and all scripts used for the analyses presented in this manuscript are available at
\url{https://github.com/DDM-Lab/greedy-hysteretic-lenient-maibl}. The codes for the comparative models are available at \url{https://github.com/gjp1203/nui_in_madrl}.

\begin{acks}
This research was partly sponsored by the Defense Advanced Research Projects Agency and was accomplished under Grant Number W911NF-20-1-0006 and by AFRL Award FA8650-20-F-6212 subaward number 1990692 to Cleotilde Gonzalez.
\end{acks}

%\bibliographystyle{ACM-Reference-Format}
%\bibliography{cas-refs}


% \appendix









\end{document}